\documentclass[letterpaper, 10 pt, conference]{ieeeconf}  

\IEEEoverridecommandlockouts                              

\usepackage{hyperref}
\usepackage{url}
\usepackage{makecell}
\usepackage{graphicx}
\usepackage{multirow}
\usepackage{placeins}
\usepackage{rotating}
\usepackage{amsmath,amsfonts,bm}
\usepackage{lscape}
\usepackage{subcaption}
\usepackage{xcolor}

\title{\LARGE \bf
Benchmarking Smoothness and Reducing High-Frequency Oscillations in Continuous Control Policies
}

\author{Guilherme Christmann*, Ying-Sheng Luo*, Hanjaya Mandala*, and Wei-Chao Chen \\
Inventec Corporation, Taipei, Taiwan \\ 
\textit{\{guilherme.christmann, luo.ying-sheng, hsu.hanjaya, chen.wei-chao\}@inventec.com}
\thanks{*These authors contributed equally, listed alphabetically by last name.}
}

\begin{document}

\maketitle

\thispagestyle{empty}
\pagestyle{empty}

\begin{abstract}

Reinforcement learning (RL) policies are prone to high-frequency oscillations, especially undesirable when deploying to hardware in the real-world. In this paper, we identify, categorize, and compare methods from the literature that aim to mitigate high-frequency oscillations in deep RL. We define two broad classes: loss regularization and architectural methods. At their core, these methods incentivize learning a smooth mapping, such that nearby states in the input space produce nearby actions in the output space. We present benchmarks in terms of policy performance and control smoothness on traditional RL environments from the Gymnasium and a complex manipulation task, as well as three robotics locomotion tasks that include deployment and evaluation with real-world hardware. Finally, we also propose hybrid methods that combine elements from both loss regularization and architectural methods. We find that the best-performing hybrid outperforms other methods, and improves control smoothness by $26.8$\% over the baseline, with a worst-case performance degradation of just $2.8$\%.

\end{abstract}

\section{Introduction}
    \par Reinforcement learning (RL) policies are prone to high-frequency oscillations. When no limitations or constraints are imposed in either the learning or in the environment, RL agents commonly develop exploitative behavior that maximizes reward to the detriment of everything else. Chasing high task performance (reward) is the goal of learning, but there are scenarios where additional factors must be considered. For example, when deploying a policy to hardware in the real-world high-frequency oscillations are especially undesirable as they can cause damage to the actuators and other hardware.
    
    \par A straightforward way to mitigate the issue is to include penalization terms as part of the reward function. However, the learning algorithm's tendency to exploit the reward function can lead to policies where subpar performance is preferred in favor of smoothness. Furthermore, reward function design is a complex matter, and can be difficult to express for many tasks in the first place \cite{gupta2022unpacking, laud2003influence, eschmann2021reward}. Adding additional penalization terms for high-frequency oscillations essentially modifies the original learning objective, and can be difficult to tune. If the penalization weight is too large, the policy might prefer to not do much at all to avoid large negative rewards. On the other hand, if the weight is too small it might choose to ignore it and still generate high-frequency oscillations. An ideal method should allow us to maintain the originally designed reward function and remove the need to add new elements of complexity.

    \begin{figure}[t!]
        \centering
        \includegraphics[width=0.48\textwidth]{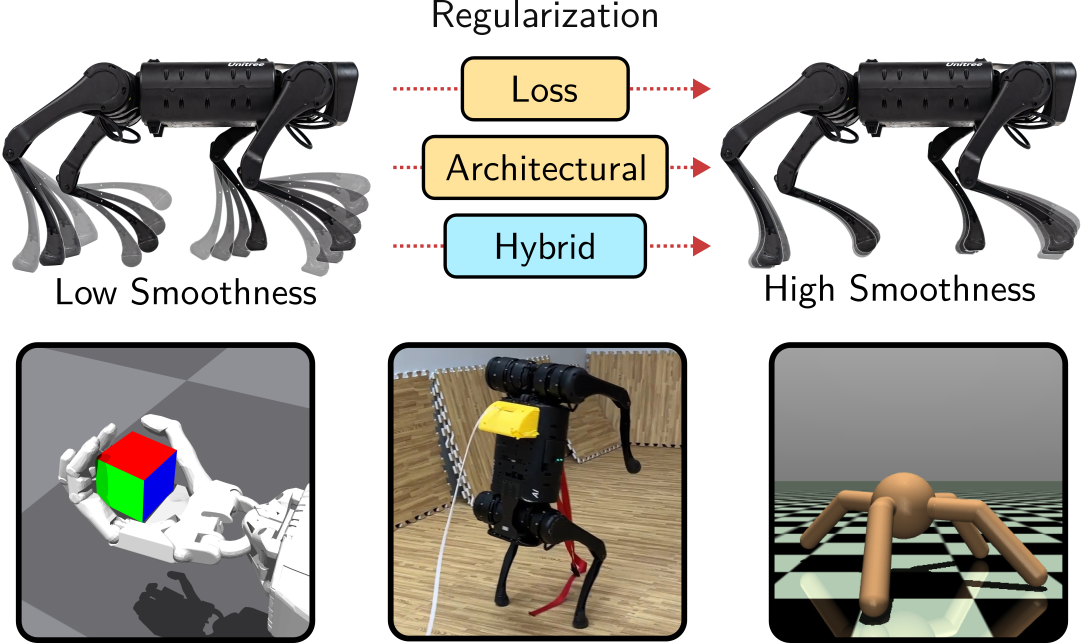}
        \caption{We investigate the use of different classes of regularization to produce smooth control policies in several simulation and real-world environments.}
        \label{fig:teaser}
    \end{figure}

    \par Another approach to reducing high-frequency oscillations is to filter the actions outputted from the policy, for example with a low-pass filter. In terms of the classic agent-environment diagram in RL \cite{sutton2018reinforcement}, this type of approach can be construed as adding a constraint to the environment rather than to the agent (or policy) itself. In fact, filtering the actions can lead to even larger oscillations in the raw outputs of the policy. That being said, filtering is effective and quite common in practice, especially in robotics applications \cite{peng2020learning}. A major drawback of using a traditional filter, e.g. a low-pass filter, is that it has memory. This means that if the observation space does not include past actions and past observations the policy will not be able to learn an effective model, as it violates the assumption of a Markov Decision Process \cite{van2012reinforcement}. Although this can be solved by keeping a history buffer over multiple steps it requires larger models in terms of parameters and complexity \cite{mysore2021regularizing}. 
    
    \par In this work, we categorize, adapt, and compare methods that aim to mitigate the problem of high-frequency oscillations in deep RL (Figure~\ref{fig:teaser}). We focus on methods that do not rely on explicit reward penalization terms, or environment modifications such as post-processing actions. Rather, we identify two classes of methods in the literature: loss regularization and architectural methods. At their core, these methods incentivize or impose constraints such that the policy learns to produce smooth mappings between input and output space, without modifying the reward function design. A mapping can be considered smooth when states that are nearby in the input space produce nearby actions in the output space \cite{shen2020deep}. A common mechanism employed by multiple existing works is to constrain the upper bound of the Lipschitz constant of the policy network \cite{fazlyab2019efficient}, either globally \cite{liu2022learning} or in a local manner \cite{kobayashi2022l2c2, song2023lipsnet}. However, we wish to note that in this work we are not concerned with the actual smoothness of the mapping itself, but rather on the observable smoothness in the form of action oscillations during test time. 
    
    \par We benchmark several methods in terms of policy performance and action smoothness. For traditional RL methods we used Gymnasium \cite{towers_gymnasium_2023} and for complex manipulation and locomotion we used Isaac Gym \cite{makoviychuk2021isaac}. Additionally, for the three locomotion tasks we include evaluation and deployment of every method in the real-world. Our paper also proposes novel hybrid methods that outperform the existing approaches in terms of smoothness and task performance trade-offs. Our contributions can be summarized as follows:

    \begin{itemize}
        \item Categorizing existing methods in two broad classes: loss regularization methods and architectural methods;
        \item Benchmarking and directly comparing action smoothing methods in classical RL simulation as well as application-focused and complex deployment scenarios in the real-world;
        \item Proposing novel hybrid methods that combine elements from other existing methods and outperform existing ones.
    \end{itemize}

\section{Related Works and Methods Categorization}
\label{sec:related-works}

    \par \textbf{Benchmarking in RL}. Reinforcement learning is a diverse field with diverse tasks and algorithms. With such a large array of possibilities, it is common for practitioners to look for benchmarks to aid in algorithm selection. \cite{duan2016benchmarking} presented a benchmark for continuous control policies in several classic tasks as well as more complex tasks such as humanoid locomotion in 3D simulation. Other works have performed benchmarks that focus on domains such as meta reinforcement learning \cite{yu2020meta}, manipulation tasks \cite{fan2018surreal}, and real-world deployment \cite{mahmood2018benchmarking, gurtler2023benchmarking}. In the context of smooth policies, past works have presented brief comparisons as a way to validate their method \cite{song2023lipsnet, kobayashi2022l2c2, mysore2021regularizing}. However, a comprehensive study that considers a holistic approach does not currently exist.
    
    \par In our work, we aim to fill this gap in the literature with a comprehensive comparison among methods that learn smooth policies without modifying the original reward function. We identify two major method classes: loss regularization and architectural methods. In the remainder of this section, we describe the general form and characteristics of each class followed by specific methods of that class. We also describe novel hybrid methods that combine elements from both loss regularization and architectural approaches.

\subsection{Loss Regularization Methods}

    \par Loss regularization methods aim to reduce the action oscillation frequency by adding regularization components to the standard RL loss objective, rather than directly in the reward function. They have the general form of

    \begin{equation}
        \mathcal{L} = \mathcal{L}_{\text{RL}} + \mathcal{L}_{Reg}
    \end{equation}

    \noindent where $\mathcal{L}_{\text{RL}}$ is a policy gradient loss such as PPO~\cite{schulman2017proximal}, TRPO~\cite{schulman2015trust}, and similar methods; and $\mathcal{L}_{Reg}$ is the regularization loss. We investigate two recent methods from the literature that fit the definition of the loss regularization class.

    \par \textbf{CAPS \cite{mysore2021regularizing}}. Uses two regularization components. The first is a temporal component $\mathcal{L}_T$, which minimizes the distance between the actions of two consecutive states $s_t$ and $s_{t+1}$. The second is a spatial component $\mathcal{L}_S$ that minimizes the difference between the state $s_t$ and a state $\bar s_t$ sampled from a normal distribution in the neighborhood of $s_t$. This takes the form of

    \begin{align}
        \label{eq:caps}
        \mathcal{L}_T &= D(\pi_\theta(s_t), \pi_\theta(s_{t+1})) \nonumber \\
        \mathcal{L}_S &= D(\pi_\theta(s_t), \pi_\theta(\bar s_t)),\quad \text{where}\ \bar s_t \sim \mathcal{N}(s_t, \sigma) \nonumber \\
        \mathcal{L}_\text{CAPS} &= \lambda_T \mathcal{L}_T + \lambda_S \mathcal{L}_S
    \end{align}

    \noindent where $\pi_\theta$ is the actor network, $D(\cdot)$ is a distance function, and $\lambda_T$, $\lambda_S$, and $\sigma$ are hyperparameters to be tuned. This method is similar to the one proposed by \cite{shen2020deep}, with the main distinction that CAPS uses the L2 distance between sampled actions, while \cite{shen2020deep} employed KL divergence on the output distributions.

    \par For scenarios in our work that overlap with the original we use the same hyperparameters from the original paper \cite{mysore2021regularizing}. For the new locomotion and manipulation environments only present in our work, we performed a short hyperparameter search and chose the best values.
    
    \par \textbf{L2C2 \cite{kobayashi2022l2c2}}. Uses two regularization components with a similar mechanism to the spatial component from CAPS. Distinctively, the regularization is employed both to the outputs of the actor-network $\pi_\theta$ and the value network $V_\theta$. Additionally, the sampling distance is bounded relative to the distance of two consecutive states $s_t$ and $s_{t+1}$, rather than a predefined hyperparameter as in CAPS. The L2C2 regularization is computed in the following way

    \begin{align}
        \bar s_t &= s_t + (s_{t+1} - s_t) \cdot u,\quad  \text{where}\ u \sim \mathcal{U(.)} \nonumber \\
        \mathcal{L}_{s, \pi} &= D(\pi_\theta(s_t), \pi_\theta(\bar s_t)) \nonumber \\
        \mathcal{L}_{s, V} &= D(V_\theta(s_t), V_\theta(\bar s_t)) \nonumber \\
        \mathcal{L}_\text{L2C2} &= \lambda_\pi \mathcal{L}_{s, \pi} + \lambda_V \mathcal{L}_{s, V}
    \end{align}

    \noindent where $\mathcal{U}$ is a uniform distribution, $D$ is a distance metric, $\pi_\theta$ and $V_\theta$ are the actor and value network, and $\lambda_\pi$ and $\lambda_V$ are weights for each regularization component. For brevity, the uniform sampling details and its hyperparameters are omitted here. We invite the reader to read the original work from \cite{kobayashi2022l2c2} for an in-depth discussion of the state sampling and definition of the hyperparameters.

    \par \textbf{L2C2} and \textbf{CAPS} are similar, with the main difference being the sampling method. It could be argued that the temporal element of \textbf{CAPS} is redundant since a state that is sampled nearby and two consecutive states should produce more or less the same regularization signal. As such, \textbf{L2C2} drops the temporal element in favor of optimizing both the actor and the value network outputs with a spatial regularization.

\subsection{Architectural Methods}
    \par Architectural methods aim to reduce the oscillation frequency of the actions by modifying the learning components of the network. In the case of the Lipschitz based methods \cite{liu2022learning, song2023lipsnet} they also add an element to the loss function. However, the objective function is used to constrain the upper bound of the Lipschitz value of the network, rather than directly optimizing state-action differences as in the loss regularization category.

    \par \textbf{Spectral Normalization -- Local SN \cite{takase2022stability}}. Spectral normalization is most commonly used to stabilize the training of Generative Adversarial Networks \cite{miyato2018spectral}. It consists of a rescaling operation applied to the weights of a layer by its spectral norm $\sigma(\bm{W})$. The normalized weights are given by $\bm{W}_{SN} = \delta \cdot \frac{\bm{W}}{\sigma(\bm{W})}$. In the context of reinforcement learning, past works have proposed global and local variants of the spectral normalization \cite{takase2022stability}. The difference between the global and local variants is that spectral normalization is applied to every layer in the global version, and only to the output layer in the local version. In our work, we investigate the local variant \textbf{Local SN} due to its significantly better performance reported by the original authors \cite{takase2022stability}.

    \par We implement this method using the spectral normalization existent in \textit{PyTorch}. Our implementation is equivalent to the original description in \cite{takase2022stability} with a $\delta=1.0$. This method does not have any other hyperparameters.

    \par \textbf{Liu-Lipschitz \cite{liu2022learning}}. This approach was originally used to learn a smooth mapping for neural distance field networks, such that interpolation and extrapolation of shapes is possible. The method constrains the Lipschitz upper bound of the network, as a learnable parameter $c_i$ per layer. The weights of each network layer are normalized with regards to $c_i$ and the layer's outputs are computed as such

    \begin{align}
        y &= \sigma(\hat{W_i} \cdot x + b_i) \nonumber \\ 
        \hat{W_i} &= normalization(W_i, softplus(c_i))
    \end{align}

    \noindent where $\hat{W_i}$ are the normalized weights and $\sigma(.)$ is an activation function. For brevity, we omit the implementation details of the normalization procedure and invite the reader to verify the original work \cite{liu2022learning}. This method also includes a loss function element that minimizes the values of $c_i$ and has the form

    \begin{equation}
        \mathcal{L}_c = \lambda \prod_{i}^{N} softplus(c_i)
    \end{equation}

    \noindent where $\lambda$ is a tunable hyperparameter, and $N$ is the number of layers in the network, with a single $c_i$ per layer.
    
    \par \textbf{LipsNet \cite{song2023lipsnet}}. The most recent out of all the methods investigated. It proposes a novel network module called \textbf{LipsNet} that can be used as plug and play replacement for a traditional feedforward layer. Specifically, we investigate the best-performing variant \textbf{LipsNet-L}, whose output is computed as such
    \begin{equation}
        y = K(x) \cdot \frac{f(x)}{||\nabla f(x)|| + \epsilon},
    \end{equation}
    where $f(x)$ is a conventional feedforward layer and $||\nabla f(x)||$ is the 2-norm of the Jacobian matrix relative to the input $x$, $K(x)$ is the Lipschitz value modeled by a feedforward network $K$ conditioned on the input $x$, and $\epsilon$ is a small positive value to avoid division by zero. 
    
    \par The authors of \textbf{LipsNet} provided an open-source implementation of their method. However, we noted a few differences from the original description in their work. Specifically, their paper \cite{song2023lipsnet} states that the activation of the $K(x)$ module is a softplus function, but in the open-source code a linear activation was used. In our implementation, we used a softplus activation as described in the original paper\footnote{We have contacted the authors of \textbf{LipsNet} and verified that our implementation as described in this work indeed reflects the original one used in their experiments. The open-source implementation has since been corrected to use a softplus activation.}. Additionally, we opted to not use a $tanh$ squashing function in the outputs of the network and instead use a linear activation, the same as every other method we experiment with in this work. 
    
\section{Method and Experimental Setup}

    \par All experiments are run using PPO with a focus on continuous observations and continuous action spaces. We benchmark traditional RL environments using Gymnasium~\cite{towers_gymnasium_2023}, and robotics application scenarios with Isaac Gym~\cite{makoviychuk2021isaac} and sim2real deployment to real hardware \cite{peng2018sim, zhao2020sim}. The base PPO implementations used are \textit{Stable Baselines}~\cite{stable-baselines} for the Gymnasium environments and the \textit{RL Games}~\cite{rl-games2021} implementation for Isaac Gym. We extended these implementations with support for all the methods outlined in Section~\ref{sec:related-works}. All of our implementations are written using \textit{PyTorch}.
    
    \par For every environment and every method, we trained policies from scratch using 9 different random seeds. Where applicable, we utilized the same hyperparameters for the same environments presented in the original works. In other cases, we tuned the parameters for better performance. The complete hyperparameters used in our investigation are presented in Table~\ref{tab:hyperparams_all_methods}.

    \begin{table*}[ht]
\vspace{0.6em}
\centering
\begin{tabular}{|l|l|l|l|l|l|l|}
\hline
\textbf{Method}                 & \textbf{Parameter}                         & \textbf{Gymnasium} & \textbf{ShadowHand} & \textbf{Motion Imitation} & \textbf{Velocity} & \textbf{Handstand} \\ \hline
\multirow{3}{*}{CAPS}           & $\sigma$                                   & 0.1                & 0.2                   & 0.2                       & 0.2                          & 0.2                  \\ \cline{2-7} 
                                & $\lambda_T$                                & 0.1                & 0.01                   & 0.01                      & 0.01                         & 0.01                  \\ \cline{2-7} 
                                & $\lambda_S$                                & 0.5                & 0.05                   & 0.05                      & 0.05                         & 0.05                  \\ \hline
\multirow{3}{*}{L2C2}           & $\sigma$                                   & 1.0                & 1.0                   & 1.0                       & 1.0                            & 1.0                  \\ \cline{2-7} 
                                & $\underline{\lambda}$                      & 0.0                & 0.01                   & 0.01                      & 0.01                            & 0.01                  \\ \cline{2-7}
                                & $\overline{\lambda}$                       & 1.0                & 1.0                   & 1.0                       & 1.0                            & 1.0                  \\ \cline{2-7}
                                & $\beta$                                    & 0.1                & 0.1                   & 0.1                       & 0.1                            & 0.1                  \\ \hline
\multirow{7}{*}{LipsNet}        & Weight $\lambda$                           & 0.1                & 0.00001                   & 0.001                     & 0.001                          & 0.0001                  \\ \cline{2-7} 
                                & $\epsilon$                                 & 0.0001             & 0.0001                   & 0.0001                    & 0.0001                       & 0.0001                  \\ \cline{2-7} 
                                & Initial Lipschitz constant $K_\text{init}$ & 1.0                & 1.0                   & 1.0                       & 1.0                          & 1.0                  \\ \cline{2-7} 
                                & Hidden layers in $f(x)$                    & {[}64, 64{]}       & {[}512, 256{]}                   & {[}512, 256{]}            & {[}512, 256{]}                 & {[}512, 256, 128{]}                  \\ \cline{2-7} 
                                & Activation in $f(x)$                       & ELU                & ELU                   & ELU                       & ELU                          & ELU                  \\ \cline{2-7} 
                                & Hidden layers in $K(x)$                    & {[}32{]}           & {[}32{]}                   & {[}32{]}                  & {[}32{]}                     & {[}32{]}                  \\ \cline{2-7} 
                                & Activation in $K(x)$                       & Tanh               & Tanh                   & Tanh                      & Tanh                         & Tanh                  \\ \hline
\multirow{2}{*}{Liu-Lipschitz}  & Weight $\lambda$                           & $1 \times 10^{-6}$ & $1 \times 10^{-7}$                   & $1 \times 10^{-6}$        & $1 \times 10^{-5}$           & $1 \times 10^{-6}$                  \\ \cline{2-7} 
                                & Initial Lipschitz constant                 & 10.0               & 10.0                   & 10.0                      & 1.0                         & 10.0                  \\ \hline

\end{tabular}
\caption{Hyperparameters used during training for every method.}
\label{tab:hyperparams_all_methods}
\end{table*}

\subsection{Hybrid Methods}

    \par We investigate the effectiveness of hybrid methods that combine elements from architectural as well as loss regularization approaches. Specifically, we focus on the combination of a \textbf{LipsNet} style network with additional regularization components in the style of \textbf{L2C2} and \textbf{CAPS}. We exclude \textbf{Local SN} due to inferior performance and inferior training stability and exclude \textbf{Liu-Lipschitz} due to the method similarity with \textbf{LipsNet} but inferior performance. We propose, experiment, and analyze two novel hybrid methods: \textbf{LipsNet~+~CAPS}, and \textbf{LipsNet~+~L2C2}.
    
\subsection{Metrics}
    \par \textbf{Cumulative Return}. The cumulative sum of the reward at every step throughout a whole episode $C = \sum_{t=0}^N R_t$. It provides a measure of the task performance of the policy. This metric is environment dependent and is used primarily to analyze the trade-off between smoothness and performance.

    \par \textbf{Smoothness}. We adopt the same smoothness metric as \cite{mysore2021regularizing}, computed from the frequency spectrum of a Fast Fourier Transform (FFT). The smoothness measure $Sm$ computes a normalized weighted mean frequency and has the form

    \begin{equation}
        Sm = \frac{2}{n \ f_s} \sum_{i=1}^{n}M_i f_i,
    \end{equation}

    \noindent where $n$ is the number of frequency bands, $f_s$ the sampling frequency, and, $M_i$ and $f_i$ are the amplitude and frequency of band $i$, respectively. Higher values indicate the presence of high-frequency components of large magnitude, and lower values indicate a smoother control signal. In the same manner as the cumulative return, a good smoothness value differs from environment to environment but is independent of the policy control frequency.

\subsection{Evaluation Scenarios}
    \par \textbf{Gymnasium Baselines}. Gymnasium \cite{towers_gymnasium_2023} provides standard and classical RL environments for easy and diverse comparisons across different algorithms. We use it to evaluate 4 classic continuous control environments: Pendulum-v1, Reacher-v4, LunarLander-v2 (Continuous version), and Ant-v4. Because Pendulum-v1 is a simpler environment we train the policies for just $150$k timesteps, while the remaining environments train for a total of $400$k timesteps. For evaluation, the metrics are computed from 1000 independent episodes for each training seed and averaged.

    \par \textbf{Robotics Applications}. With Isaac Gym \cite{makoviychuk2021isaac} we train policies to execute three locomotion tasks with a quadruped robot and a manipulation task with a highly dexterous hand. The manipulation task \textit{ShadowHand} is one of the standard tasks bundled with Isaac Gym and consists of manipulating a cube to match a target orientation \cite{andrychowicz2020learning}. The other three tasks are implemented by us with additional deployment on real-world hardware. \textit{Motion Imitation}, where the agent is rewarded for matching the states of a motion-captured pacing animation \cite{peng2018deepmimic, peng2020learning, icra2023ch}; \textit{Velocity} where the agent is rewarded for matching a velocity vector \cite{rudin2021learning}. Locomotion emerges as the result of tracking the velocity command plus additional regularization terms. Note that we do not use an action penalization term in the reward design of this task as was done in past works \cite{rudin2021learning}; and \textit{Handstand}, where the agent is rewarded for standing on its hind legs and maintaining an upright orientation by tracking a target orientation vector. The task includes additional reward regularization terms to minimize joint changes as well as linear and angular velocities. \textit{Motion Imitation} and \textit{Handstand} task are trained for $150$M timesteps, and \textit{Velocity} and \textit{ShadowHand} are trained for $300$M timesteps. For each training seed, the evaluation metrics are collected and averaged from $10$k trajectories for \textit{Motion Imitation} and \textit{ShadowHand} and $50$k trajectories for \textit{Velocity} and \textit{Handstand}.
    
    \begin{table}[ht!]
    \centering
    \begin{tabular}{ |l|c|c| } 
        \hline
        \textbf{Parameter} & \textbf{Value} & \textbf{Type} \\
        \hline
        Action Noise                    & 0.02 & Additive \\ 
        Rigid Bodies Mass               & [0.95, 1.05] & Scaling \\
        Stiffness Gain (PD Controller)  & [-10\%, +10\%] & -- \\
        Damping Gain (PD Controller)    & [-15\%, +15\%] & -- \\
        Ground Friction                 & [0.1, 1.5] & -- \\
        Sensor Noise - Orientation      & 0.06 & Additive \\
        Sensor Noise - Linear Velocity  & 0.25 & Additive \\
        Sensor Noise - Angular Velocity & 0.3 & Additive \\
        Sensor Noise - Joint Angles     & 0.02 & Additive \\
        Sensor Noise - Feet Contacts    & 0.2 & Probability \\
        \hline
    \end{tabular}
    \caption{Domain randomization parameters used to train the policies that are deployed to the real-world.}
    \label{tab:domain_rand_params}
\end{table}

    \begin{figure}[ht!]
  \vspace{0.6em}
  \centering
  \includegraphics[width=0.99\columnwidth]{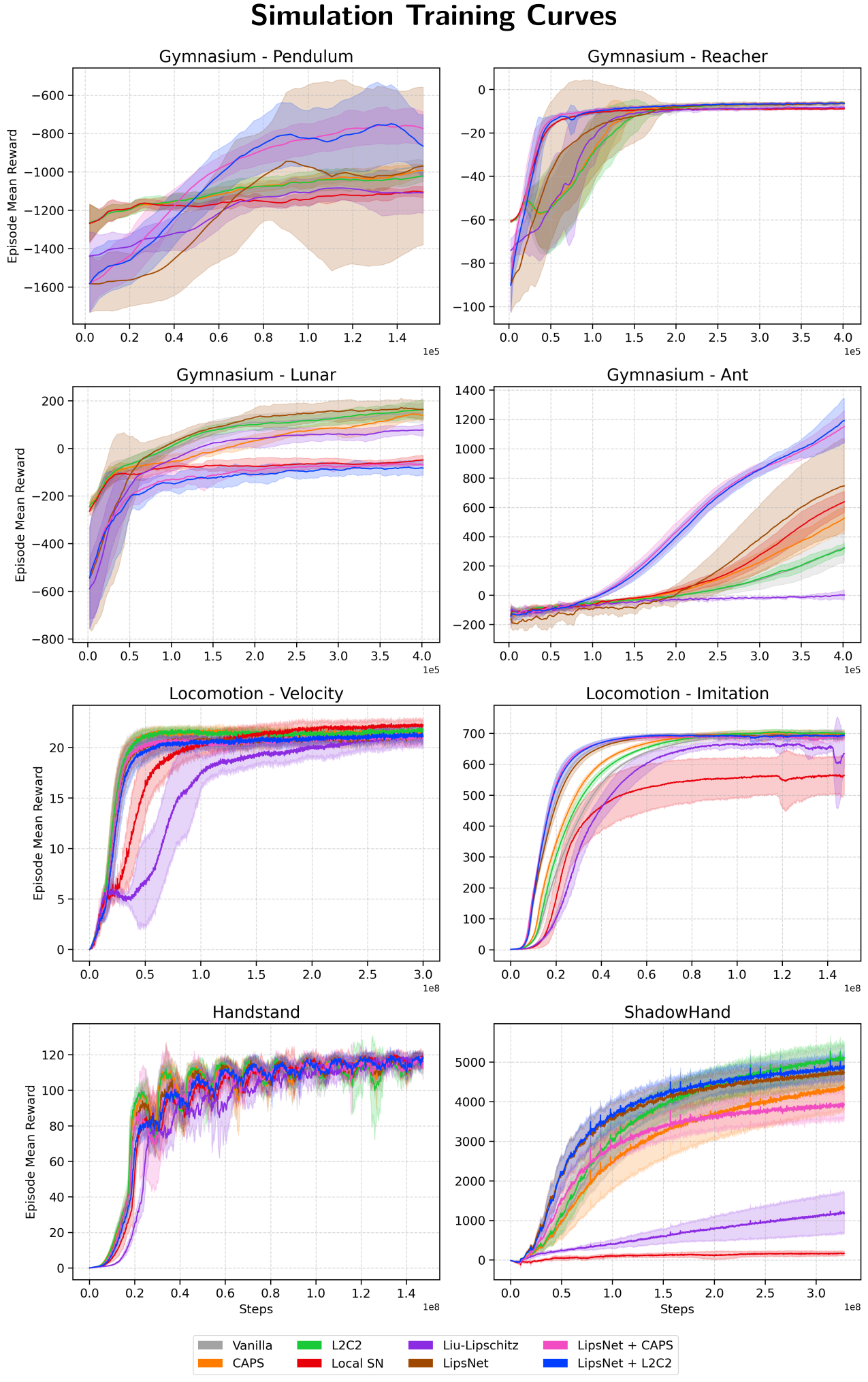}
  \caption{Reward curves during training for 9 seeds. The hybrid methods \textbf{LipsNet + CAPS} and \textbf{LipsNet + L2C2} show superior or comparable all environments, except in \textit{Lunar}. }
  \label{fig:training_curves}
\end{figure}

    \par \textbf{Real-World Experiments}. For the three locomotion tasks described above, we conduct real-world experiments with a quadruped robot. The policies are trained with domain randomization (DR) \cite{tobin2017domain} to ensure a successful sim-to-real transfer. By adding noise to elements of the simulation the policy learns to perform reasonably across a larger distribution of states. The simulation parameters randomized during training of the deployment version of the policies are presented in Table~\ref{tab:domain_rand_params}. We investigate the effect of every single method in the real-world, with an additional ablation case where a vanilla policy is trained without DR. This case demonstrates that the use of DR already produces smoother control policies. The evaluation metrics at deployment time in the real-world are computed from 6-second trajectories recorded during the execution of the policies. We compute the smoothness $Sm$ of the whole 6-second trajectory, and the cumulative return is the accumulated reward value from each step throughout the whole trajectory. Note that for the Handstand task, we do not deploy a Vanilla policy without domain randomization, due to the risk of hardware damage from bad policy performance and large oscillations.

    \begin{figure}[ht!]
    \vspace{0.6em}
    \centering
    \includegraphics[width=0.48\textwidth]{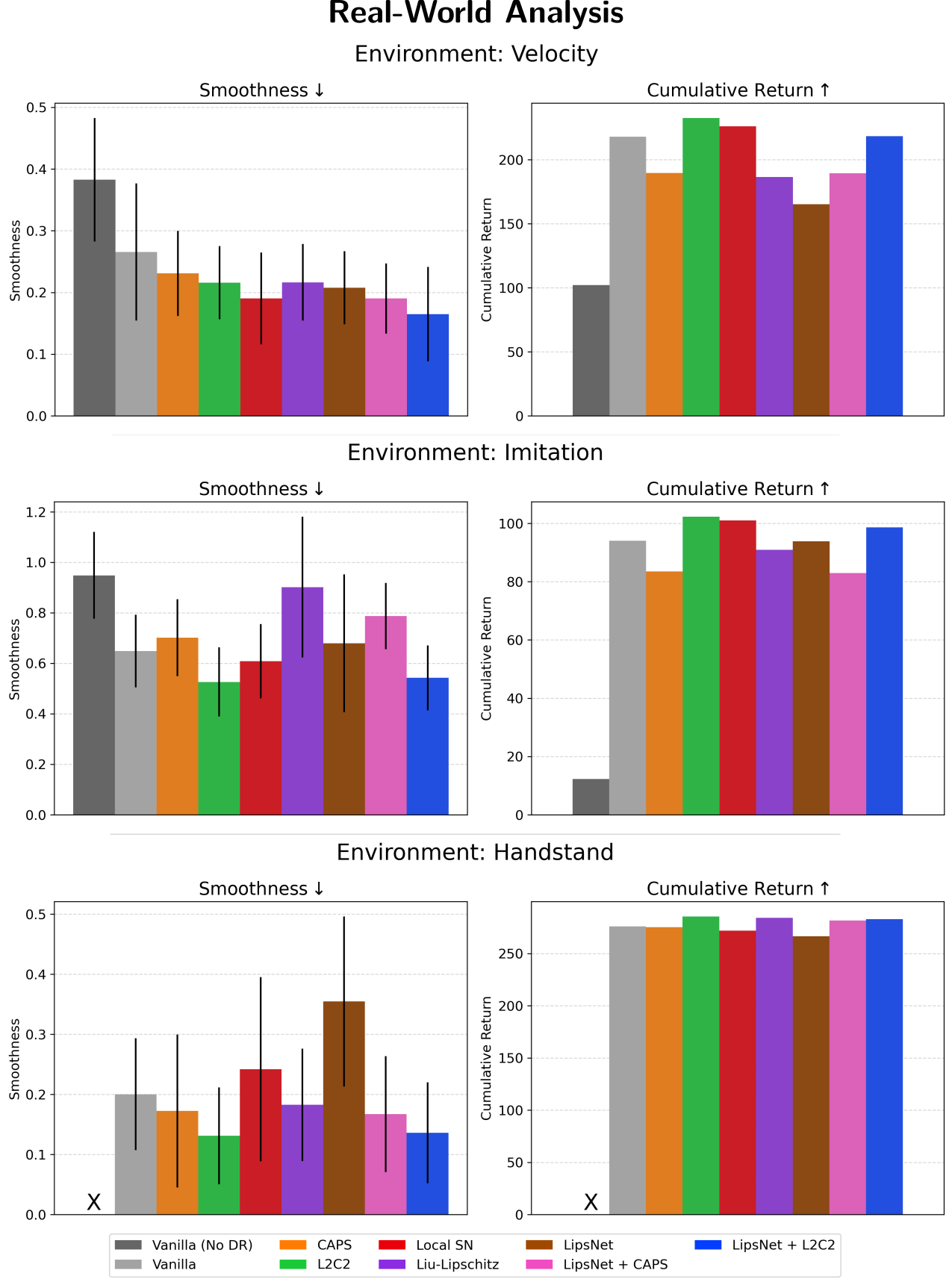}
    \caption{The methods are evaluated with the real-world robot. Every method achieves similar task performance (measured as cumulative return). The hybrid methods consistently outperformed other methods in regards to smoothness compared to the other methods.}
    \label{fig:real-world-results}
\end{figure}

\section{Experiments and Results}

\par We benchmarked a total of 7 distinct approaches plus a vanilla baseline, across 8 different scenarios, including classic RL environments and complex robotic simulations with additional deployment in the real-world. The complete results of our simulation benchmark are consolidated in Table~\ref{tab:results_table}. Additionally, for sample efficiency analysis, the training curves showcasing the episode mean reward over time are depicted in Figure~\ref{fig:training_curves}. 

\begin{table*}[ht!]
\vspace{0.6em}
\centering
\begin{tabular}{|l|llll|}
\multicolumn{5}{c}{\textbf{Cumulative Return} $\uparrow$} \\
\hline

Environment                                                                  & Pendulum                & Ant                    & Reacher                 & Lunar              \\
\hline
\textbf{Vanilla}                                                     & $ -944 \pm 57 $       & $ 833 \pm 110 $      & $ -6.05 \pm 0.46 $      & $ 170 \pm 49 $     \\
\textbf{CAPS} -- \scriptsize{\cite{mysore2021regularizing}}         & $ -940 \pm 51 $       & $ 1027 \pm 135 $     & $ \bm{-5.98 \pm 0.21} $ & $ -117 \pm 43 $    \\
\textbf{L2C2} -- \scriptsize{\cite{kobayashi2022l2c2}}              & $ -962 \pm 44 $       & $ 791 \pm 104 $      & $ -6.14 \pm 0.49 $      & $ \bm{192 \pm 32} $\\
\textbf{Local SN} -- \scriptsize{\cite{takase2022stability}}        & $ -1099 \pm 47 $      & $ 1108 \pm 174 $     & $ -8.73 \pm 0.31 $      & $ -126 \pm 28 $    \\
\textbf{Liu-Lipschitz} -- \scriptsize{\cite{liu2022learning}}       & $ -1056 \pm 111 $     & $ 137 \pm 210 $      & $ -7.68 \pm 0.86 $      & $ 92 \pm 70 $      \\
\textbf{LipsNet}           -- \scriptsize{\cite{song2023lipsnet}}   & $ -934 \pm 445 $      & $ 959 \pm 506 $      & $ -6.34 \pm 0.69 $      & $ 114 \pm 71 $     \\
\textbf{LipsNet + CAPS} -- \scriptsize{Hybrid}                	     & $ \bm{-737 \pm 182} $ & $ \bm{1683 \pm 228}$ & $ -6.13 \pm 0.30 $      & $ -304 \pm 22 $    \\
\textbf{LipsNet + L2C2}    -- \scriptsize{Hybrid}             	     & $ -870 \pm 172 $      & $ 1684 \pm 415 $     & $ -6.27 \pm 0.44 $      & $ -281 \pm 72 $    \\
\hline
Environment                                                          &  ShadowHand           & Imitation             & Velocity               & Handstand          \\
\hline                                                                                         
\textbf{Vanilla}                                                     & $ 5025 \pm 460 $      & $ 697 \pm 19 $        & $ 5.98 \pm 0.05 $      & $ 3.37 \pm 0.10 $      \\
\textbf{CAPS} -- \scriptsize{\cite{mysore2021regularizing}}         & $ 4421 \pm 576 $      & $ 689 \pm 26 $        & $ \bm{5.99 \pm 0.03} $ & $ 3.35 \pm 0.06 $      \\
\textbf{L2C2} -- \scriptsize{\cite{kobayashi2022l2c2}}              & $ \bm{5190 \pm 390} $      & $ \bm{697 \pm 17} $   & $ 5.86 \pm 0.04 $      & $ 3.41 \pm 0.06 $      \\
\textbf{Local SN} -- \scriptsize{\cite{takase2022stability}}        & $ 166 \pm 93 $        & $ 522 \pm 132 $       & $ 5.71 \pm 0.19 $      & $ 3.37 \pm 0.07 $      \\
\textbf{Liu-Lipschitz} -- \scriptsize{\cite{liu2022learning}}       & $ 1213 \pm 538 $       & $ 644 \pm 53 $        & $ 5.34 \pm 0.12 $      & $ 3.38 \pm 0.03 $      \\
\textbf{LipsNet}           -- \scriptsize{\cite{song2023lipsnet}}   & $ 4784 \pm 333 $      & $ 682 \pm 27 $        & $ 5.86 \pm 0.12 $      & $ 3.40 \pm 0.06 $      \\
\textbf{LipsNet + CAPS} -- \scriptsize{Hybrid}                       & $ 3923 \pm 347 $      & $ 673 \pm 27 $        & $ 5.91 \pm 0.06 $      & $ 3.40 \pm 0.06 $      \\
\textbf{LipsNet + L2C2}    -- \scriptsize{Hybrid}                    & $ 4913 \pm 305 $      & $ 678 \pm 33 $        & $ 5.83 \pm 0.13 $      & $ \bm{3.45 \pm 0.05} $ \\
\hline
\multicolumn{5}{c}{\textbf{Smoothness $Sm$ $\downarrow$}} \\
\hline
Environment                                                          & Pendulum                & Ant                    & Reacher $\cdot 10^1$       & Lunar $\cdot 10^1$             \\
\hline
\textbf{Vanilla}                                                     & $ 0.77 \pm 0.04 $       & $ 1.94 \pm 0.57 $      & $ 0.62 \pm 1.08 $      & $ 6.24 \pm 7.05 $     \\
\textbf{CAPS} -- \scriptsize{\cite{mysore2021regularizing}}         & $ 0.73 \pm 0.06 $       & $ 1.11 \pm 0.40 $      & $ 0.52 \pm 0.05 $      & $ 2.20 \pm 0.66 $    \\
\textbf{L2C2} -- \scriptsize{\cite{kobayashi2022l2c2}}              & $ 0.73 \pm 0.08 $       & $ 1.49 \pm 0.43 $      & $ 0.56 \pm 0.09 $      & $ 5.43 \pm 0.78 $\\
\textbf{Local SN} -- \scriptsize{\cite{takase2022stability}}        & $ 0.40 \pm 0.07 $       & $ \bm{0.70 \pm 0.29} $ & $ 0.60 \pm 0.11 $      & $ 3.90 \pm 0.36 $    \\
\textbf{Liu-Lipschitz} -- \scriptsize{\cite{liu2022learning}}       & $ 0.49 \pm 0.11 $       & $ 1.07 \pm 0.28 $      & $ 0.47 \pm 0.13 $      & $ 6.23 \pm 0.76 $      \\
\textbf{LipsNet}           -- \scriptsize{\cite{song2023lipsnet}}   & $ 0.94 \pm 0.42 $       & $ 1.38 \pm 0.36 $      & $ 1.11 \pm 1.23 $      & $ 5.50 \pm 2.96 $     \\
\textbf{LipsNet + CAPS} -- \scriptsize{Hybrid}                       & $ \bm{0.31 \pm 0.08} $  & $ 0.75 \pm 0.10 $ 	    & $ \bm{0.35 \pm 0.05} $ & $ \bm{0.59 \pm 0.11} $    \\
\textbf{LipsNet + L2C2}    -- \scriptsize{Hybrid}                    & $ 0.64 \pm 0.28 $       & $ 0.87 \pm 0.17 $ 	    & $ 0.35 \pm 0.11 $ & $ 0.95 \pm 0.21 $    \\
\hline
Environment                                                          & ShadowHand         & Imitation               & Velocity               & Handstand          \\
\hline
\textbf{Vanilla}                                                     & $ 1.82 \pm 0.09 $  & $ 0.68 \pm 0.16 $       & $ 0.40 \pm 0.01 $      & $ 0.64 \pm 0.11 $   \\
\textbf{CAPS} -- \scriptsize{\cite{mysore2021regularizing}}         & $ 1.63 \pm 0.15 $  & $ 0.70 \pm 0.16 $       & $ 0.40 \pm 0.02 $      & $ 0.63 \pm 0.07 $   \\
\textbf{L2C2} -- \scriptsize{\cite{kobayashi2022l2c2}}              & $ 1.64 \pm 0.07 $  & $ 0.52 \pm 0.15 $       & $ 0.52 \pm 0.15 $      & $ 0.56 \pm 0.05 $   \\
\textbf{Local SN} -- \scriptsize{\cite{takase2022stability}}        & $ 0.09 \pm 0.08 $  & $ 0.63 \pm 0.06 $       & $ 0.35 \pm 0.19 $      & $ 0.58 \pm 0.04 $   \\
\textbf{Liu-Lipschitz} -- \scriptsize{\cite{liu2022learning}}       & $ 1.28 \pm 0.2 $	  & $ 0.66 \pm 0.10 $      & \bm{$ 0.24 \pm 0.11 $}      & $ 0.60 \pm 0.03 $   \\
\textbf{LipsNet}           -- \scriptsize{\cite{song2023lipsnet}}   & $ 1.77 \pm 0.07 $  & $ 0.65 \pm 0.13 $       & $ 0.30 \pm 0.10 $      & $ 0.55 \pm 0.03 $   \\
\textbf{LipsNet + CAPS} -- \scriptsize{Hybrid}                       & $ \bm{1.58 \pm 0.06} $  & $ 0.60 \pm 0.12 $       & $ 0.28 \pm 0.07 $      & $ 0.56 \pm 0.06 $   \\
\textbf{LipsNet + L2C2}    -- \scriptsize{Hybrid}                    & $ \bm{1.57 \pm 0.08} $  & $ \bm{0.52 \pm 0.07} $  & $\bm{0.26 \pm 0.06} $ & $ \bm{0.46 \pm 0.04} $   \\
\hline
\end{tabular}
\caption{Benchmark of task performance and smoothness of different algorithms in the literature. Each method is trained from scratch with 9 different seeds. The table shows the mean and 1 standard deviation of smoothness and return for 9 seeds. Pendulum, Ant, Reacher and Lunar are Gymnasium environments, and Imitation and Velocity are locomotion tasks in Isaac Gym.}
\label{tab:results_table}
\end{table*}

\par From the simulation results presented in Table~\ref{tab:results_table} we can infer that every method improves smoothness compared to the \textbf{Vanilla} policy. However, this comes with a performance cost in some cases. The environments \textit{Ant} and \textit{Reacher} had a high-performance variance, with many methods performing significantly worse than the baseline. Other environments provide more consistent results and serve better for a smoothness-performance tradeoff analysis. We can observe that the loss regularization methods \textbf{CAPS} and \textbf{L2C2} perform similarly in most cases, with a decent improvement in smoothness and a small performance hit overall. On the architectural methods, \textbf{Local SN} and \textbf{Liu-Lipschitz} generated even smoother policies, at the cost of a large performance deficit. \textbf{LipsNet} is inconsistent, sometimes generating smooth policies (see \textit{Velocity} and \textit{Handstand}) and others performed significantly worse than the baseline (see \textit{Ant} and \textit{Reacher}).

%
%

\par Our hybrid methods \textbf{LipsNet + CAPS} and \textbf{LipsNet + L2C2} outperforms the existing methods in nearly every environment. The \textit{Lunar} environment is the single exception where they are clearly inferior, with better smoothness but low cumulative return. We hypothesize that situations like this might happen due to the policy ``getting stuck'' too early in optimizing for smoothness, rather than task performance. More extensive hyperparameters search and better scheduling of learning rate and loss weights could yield better outcomes. Excluding \textit{Ant} and \textit{Lunar} due to high variance across methods, we can observe that \textbf{LipsNet + CAPS} produces $28.4$\% smoother control compared to the \textbf{Vanilla} baseline, while \textbf{LipsNet + L2C2} is at a close second with a $26.8$\% average improvement. In terms of performance impact, \textbf{LipsNet + CAPS} had a worst case performance degradation of $21.9$\% in the \textit{ShadowHand} environment. On the other hand, \textbf{LipsNet + L2C2} reproduced the performance of the unregularized \textbf{Vanilla} more consistently, with a worst case degradation of only $2.8$\% in \textit{Imitation}.

\par For the three robotics locomotion tasks \textit{Imitation}, \textit{Velocity}, and \textit{Handstand} we perform deployment and analysis with a real-world quadruped robot. The complete results of the real-world deployment are presented in Figure~\ref{fig:real-world-results}, which includes the \textbf{Vanilla} baseline and the 7 methods investigated in this work, plus an additional ablation of a \textbf{Vanilla} policy trained without domain randomization. Domain randomization is commonly used as a way to improve real-world deployment performance. In this work, we also note that DR significantly improves the smoothness of the control, on top of the expected performance improvement. Analyzing the bar plots we can observe that many of the methods impact measured in simulation are diminished in the real-world with the inclusion of DR. In \textit{Velocity} we can observe smoothness improves with every method compared to the \textbf{Vanilla} baseline. However, the other two scenarios \textit{Imitation} and \textit{Handstand} have more variance, with several methods ending up less smooth than the baseline with DR. Still, \textbf{L2C2} and the hybrid \textbf{LipsNet + L2C2} are the clear superior methods in the real-world experiment, outperforming every other approach. Both methods significantly improved smoothness over \textbf{Vanilla} while maintaining comparable task performance.

\par During our real-world experiments we also observed an emergent behavior when the agent suffers from large disturbances. For example, when the robot is lifted on the air or flipped over, \textbf{Vanilla} policies tend to generate high-frequency oscillations in sudden bursts, which could risk hardware damage. On the other hand, the regularized policy \textbf{LipsNet + L2C2} elegantly stops execution until the agent is reset to the ground in an upright position. We invite the reader to check the video in the supplementary material for demonstrations of this behavior.

\section{Conclusion}

\par In this work we present a benchmark of methods that can reduce the frequency of control oscillations in policies learned with reinforcement learning. We identify 7 methods from the literature and classify them according to their mechanism. We propose two broad method classes: loss regularization and architectural methods. Loss regularization methods rely purely on adding regularization elements to the standard policy gradient loss. On the other hand, architectural methods introduce modifications of network elements such as weight normalization and specialized modules that can replace traditional feedforward layers. Additionally, we also introduce and investigate two novel hybrid methods that combine properties from both method classes.

\par Our benchmark includes 4 traditional RL environments and 4 complex robotics tasks involving manipulation and locomotion. We analyze every method regarding task performance as well as the smoothness of output actions. In general, the investigated methods perform better than the unregularized \textbf{Vanilla} baseline in every scenario, with a few exceptions. We identify \textbf{LipsNet + L2C2} as the best-performing method in simulation, with a smoothness improvement of $26.8$\% over the baseline, and a worst case task performance degradation of just $2.8$\%. For the three robotics tasks that involve locomotion, we deploy and perform analysis in the real-world. Overall, \textbf{L2C2} and hybrid method \textbf{LipsNet + L2C2} showed the best trade-off between smoothness and task performance in the real-world. As such, we recommend that practitioners concerned with oscillations train policies with one of these approaches.

\par In the future, we wish to investigate tasks with an emphasis on more diverse robotics applications. Reducing high-frequency oscillations is essential for successful sim-to-real transfer and to prevent hardware damage. As such, the community can benefit from a clearer set of guidelines on how to train policies for tasks such as locomotion, pick and place, contact rich manipulation, etc.

\FloatBarrier

\bibliographystyle{IEEEtran}
\bibliography{IEEEexample}

\end{document}